\documentclass{tlp}
\usepackage{graphicx}
\usepackage{xspace}
\usepackage{color}
\usepackage{url}
\usepackage{multirow}
\usepackage{subfig}

\newcommand{\solver}{\texttt{sunny-cp}\xspace}
\newcommand{\solverseq}{\texttt{sunny-cp-seq}\xspace}
\newcommand{\solverl}{\texttt{sunny-cp}$^-$\xspace}
\newcommand{\solverll}{\texttt{sunny-cp}$^{--}$\xspace}

\begin{document}

\title{
SUNNY-CP and the MiniZinc Challenge
}
\author[]
         {Roberto Amadini \\
         Department of Computing and Information Systems, The University of Melbourne, Australia.
         \and 
         Maurizio Gabbrielli \\
         DISI, University of Bologna, Italy /  FOCUS Research Team, INRIA, France.
         \and
         Jacopo Mauro \\
         Department of Informatics, University of Oslo, Norway.
         }


%
%
\maketitle

\begin{abstract}
In Constraint Programming (CP) a portfolio solver combines a 
variety of different constraint solvers for solving a given problem. 
This fairly recent approach enables to significantly boost the performance 
of single solvers, especially when multicore architectures are exploited.
In this work we give a brief overview of the 
portfolio solver \solver, and we discuss its performance in the 
MiniZinc Challenge---the annual international 
competition for CP solvers---where it won two  
gold medals in 2015 and 2016.

\textbf{Under consideration in Theory and Practice of Logic Programming (TPLP)}
\end{abstract}

\section{Introduction}
\label{sec:intro}
In \emph{Constraint Programming} (CP) the goal is to model and solve 
Constraint Satisfaction Problems (CSPs) and 
Constraint Optimisation Problems (COPs)~\cite{handbook}.
Solving a CSP means finding a solution that satisfies all the constraints 
of the problem, while for COPs the goal is to find an optimal solution, which 
minimises (or maximises) an objective function.

A fairly recent trend to solve combinatorial problems, based on the well-known 
\textit{Algorithm Selection} problem~\cite{rice_algorithm_selection},
consists of building portfolio 
solvers~\cite{DBLP:journals/ai/GomesS01}.
A \textit{portfolio solver} is a meta-solver that exploits a collection 
of $n > 1$ constituent solvers $s_1, \dots, s_n$. When a new, unseen problem 
comes, the portfolio solver seeks to predict and run its best solver(s) 
$s_{i_1}, \dots, s_{i_k}$ (with $1 \leq k \leq n$) for solving the problem.

Despite that plenty
of Algorithm Selection 
approaches have been proposed
\cite{selection_survey,survey-algorithm-selection,prediction-state-art},
a relatively small number of portfolio solvers have 
been 
practically adopted~\cite{lopstr}. 
In particular, only few portfolio solvers participated in CP solvers 
competitions.
The first one (for solving CSPs 
only) was CPHydra~\cite{cphydra} that in 2008 won the International CSP Solver 
Competition.\footnote{The International CSP Solver 
Competition ended in 2009.} In 2013 a portfolio solver based on 
Numberjack~\cite{numberjack} attended the \emph{MiniZinc Challenge} 
(MZNC)~\cite{DBLP:journals/aim/StuckeyFSTF14}, 
nowadays the only surviving international competition for CP 
solvers.

Between 2014 and 2016, \solver was the only portfolio solver that 
joined the MZNC. Its first, sequential version
had appreciable results in the MZNC 2014 but remained off the podium. In MZNC 
2015 and 2016 its enhanced, parallel version~\cite{sunnycp2} demonstrated 
its effectiveness by winning the gold medal in the Open category of the 
challenge. 

In this paper, after a brief overview of \solver,  we 
discuss the performance it achieved in the MiniZinc Challenges 
2014---2016 and we propose directions for future works.
The lessons we learned are: 
\begin{itemize}
 \item a portfolio solver is robust even in prohibitive scenarios, 
 like the MiniZinc Challenge, 
 characterised by small-size test sets and unreliable solvers;
 \item in a multicore setting, a parallel portfolio of 
sequential solvers appears to be more fruitful than a single, parallel 
solver;
 \item \solver can be a useful baseline 
to improve the state-of-the-art for (not only) the CP field, 
where dealing with non-reliable solvers must be properly addressed.
\end{itemize}


\section{SUNNY and SUNNY-CP}
\label{sec:sunny}
In this section we provide a high-level description of \solver, referring the 
interested reader to Amadini et al. \citeyear{sunnycp,sunnycp2} for a more 
detailed 
presentation.

\solver is an open-source CP portfolio solver.
Its first 
implementation 
was sequential~\cite{sunnycp}, while the current version exploits multicore 
architectures to run more solvers in parallel and to enable   
their cooperation via bounds sharing and restarting policies. 
To the best of our knowledge, it is currently the only parallel 
portfolio solver able to solve generic CP problems encoded in Mini\-Zinc 
language~\cite{minizinc}.

\solver is built on top of SUNNY algorithm~\cite{sunny}. Given a 
set of known problems, a 
solving timeout $T$ and a portfolio $\Pi$, SUNNY uses  
the \textit{$k$-Nearest Neighbours} ($k$-NN)
algorithm to produce a sequential schedule 
$\sigma = [(s_1, t_1), \dots, (s_k, t_n)]$ where solver $s_i \in \Pi$ has to be run for 
$t_i$ seconds and $\sum_{i = 1}^{n} t_i = T$. 
The time slots $t_i$ and the ordering of solvers $s_i$ are defined 
according 
to the average performance of the solvers of $\Pi$ on the $k$ training 
instances 
closer to 
the problem to be solved. 

For each problem $p$, 
a \textit{feature vector} is computed and the 
Euclidean distance is used to retrieve the $k$ instances in the training set 
closer to 
$p$.
In a nutshell, a feature vector is a collection of numerical attributes that 
characterise a given problem instance. \solver uses several features, 
e.g., statistics over the variables, the (global) constraints, 
the objective function (when applicable), the search heuristics. In total, 
\solver uses 95 features.\footnote{The first 
version of \solver also used graph features and dynamic features, afterwards 
removed for the sake of efficiency and portability.
For more details about \solver features, please see 
\citeNS{mzn2feat} and \url{https://github.com/CP-Unibo/mzn2feat}.}

The sequential schedule $\sigma$ 
is then parallelised 
on the $c \geq 1$ available cores by running the
first and most promising 
$c-1$ solvers in the $k$-neighbourhood on the first $c-1$ cores, while the 
remaining solvers (if any) are assigned to the last available core by linearly 
widening their allocated times to cover the time window $[0, T]$. 

The notion of ``promising solver'' depends on the context. For CSPs, the 
performance is measured only in terms 
of number of solved instances and average solving time. For COPs,
also the quality of the solutions is taken into account~\cite{paper_cp}.
We might say that \solver uses a conservative policy: first, it skims the original 
portfolio by selecting a promising subset of its solvers; second, it allocates 
to each of these solvers an amount of time proportional to their 
supposed effectiveness.

Solvers are run in parallel and a \emph{``bound-and-restart''} mechanism is used 
for enabling the bounds sharing between the running COP 
solvers~\cite{sunnycp2}. 
This allows one to use
the (sub-optimal) solutions found by a solver to narrow the search space of 
the other scheduled solvers.
If there are fewer solvers than cores, \solver simply allocates a solver per 
core.

Since \solver treats solvers as black boxes, it
can not support the 
sharing of the bounds knowledge without the solvers interruption. For 
this reason, 
a restarting threshold $T_r$ is used to decide when to stop a solver 
and 
restart it with a new bound. A running solver is stopped and 
restarted when: \textit{(i)} it has 
not found a solution in the last $T_r$ seconds; \textit{(ii)} its current best 
bound is obsolete w.r.t.~the overall best bound found by another scheduled solver.

\begin{table}[t]
\centering
\scalebox{0.94}{
\begin{oldtabular}{cc}
\hline
 \textbf{Solver(s)} & \textbf{Description} \\
 \hline
 Choco$^*$, G12/FD, Gecode, JaCoP$^{**}$, Mistral$^{**}$, OR-Tools$^*$ & Finite Domain (FD) solvers\\\hline
 Chuffed, CPX, G12/LazyFD, Opturion CPX$^*$ & Lazy Clause Generation solvers\\\hline
 G12/CBC, MZN/Gurobi & MIP solvers\\\hline
 HaifaCSP$^*$ & Proof-producing CSP Solver\\\hline
 \multirow{2}{*}{iZplus$^*$}  & CP solver using local search \\
        & and no-good techniques\\\hline
 \multirow{2}{*}{MinisatID$^*$} & Combines techniques from \\
          & SAT, SMT, CP, and ASP\\\hline
 Picat-SAT$^{**}$ & Encodes CP problems into SAT\\\hline
\end{oldtabular}}
\caption{Constituent solvers of \solver. The * symbol indicates the solvers introduced in MZNC 2015, 
while ** indicates those introduced in MZNC 2016.\label{tab:solvers}}
\end{table}

Table \ref{tab:solvers} summarises the solvers used by \solver in the MZNCs 2014--2016. 
For more details about these solvers, 
see 
\citeNS{choco}, 
\citeNS{minisatid},
\citeNS{picat-sat},
\citeNS{haifacsp}, \citeNS{mzn}, \citeNS{jacop}, \citeNS{mistral}, 
\citeNS{ortools}, \citeNS{chuffed}, \citeNS{opturion}, \citeNS{izplus}.

\section{SUNNY-CP and the MiniZinc Challenge}
\label{sec:chall}
The MiniZinc Challenge 
(MZNC)~\cite{DBLP:journals/aim/StuckeyFSTF14} is the annual 
international competition for CP solvers.
Portfolio solvers compete in the ``Open'' class of MZNC, 
where all solvers are free to use multiple threads or cores, 
and no search strategy is imposed.

The scoring system of the MZNC is based on a \emph{Borda} 
count~\cite{Chevaleyre07ashort} 
where a solver $s$ is compared against each other solver $s'$ 
over 100 problem instances---belonging to different classes---defined in the 
MiniZinc language. If $s$ gives a better 
answer than $s'$ then it scores 1 point, if it gives a worse solution it scores 
0 points. 
If $s$ and $s'$ give indistinguishable answers the scoring is based on the 
solving time.\footnote{Please refer to 
\url{http://www.minizinc.org/challenge.html} for further details.}

Until MZNC 2014, the solving timeout was 15 minutes and did not include
the MiniZinc-to-FlatZinc conversion time. Starting from the MZNC 2015 this 
time has been included, and the timeout has been extended to 20 minutes.

\subsection{MiniZinc Challenges 2013--2014}
Table \ref{tab:results1314} summarizes the Open class results in the MZNCs 
2013--2014. 
\begin{table}
\subfloat[MZNC 2013, Open category.]{
\centering
\label{tab:open13}
\scalebox{0.96}{
\begin{oldtabular}{cc}
\hline
Solver & Score  \\
\hline
OR-Tools *   & 1098.85 \\ 
\textit{Chuffed}      & \textit{1034.81} \\ 
Choco *      & 973.27  \\ 
Opturion CPX & 929.76  \\
Gecode *     & 858.24  \\ 
iZplus       & 758.47  \\ 
\textit{G12/LazyFD}   & \textit{664.44}  \\ 
Mistral      & 614.62  \\ 
\textit{MZN/Gurobi}   & \textit{589.38}  \\ 
JaCoP        & 577.08  \\ 
Fzn2smt      & 556.94  \\ 
Gecoxicals   & 512.73  \\ 
\textit{MZN/CPLEX}    & \textit{447}     \\ 
\textit{G12/FD}       & \textit{426.53}  \\ 
\textbf{Numberjack} * & \textbf{383.18} \\ 
Picat        & 363.02  \\ 
\textit{G12/CBC}      & \textit{118.69}  \\
& \\
\hline
\end{oldtabular}}}
\hfill
\subfloat[MZNC 2014, Open category.]{
\centering
\label{tab:open14}
\scalebox{0.96}{
\begin{oldtabular}{cc}
\hline
Solver & Score \\
\hline
\textit{Chuffed}      & \textit{1326.02} \\ 
OR-Tools *    & 1086.97 \\ 
Opturion CPX  & 1081.02 \\ 
\textit{\textbf{sunny-cp-seq-pre}}  & \textit{\textbf{1066.46}} \\
Choco *  & 1007.61 \\ 
iZplus *  & 996.32 \\ 
\textit{\textbf{sunny-cp-seq}}  & \textit{\textbf{968.64}} \\ 
\textit{G12/LazyFD}  & 784.28 \\ 
HaifaCSP  & 781.72 \\ 
\textit{Gecode} *  & \textit{721.48} \\ 
SICStus Prolog & 710.51 \\ 
Mistral    & 705.56 \\ 
MinisatID  & 588.74 \\ 
Picat SAT  & 588.06 \\ 
JaCoP  & 550.74 \\ 
\textit{G12/FD}    & \textit{528.26} \\ 
Picat CP  & 404.88 \\ 
Concrete  & 353.74 \\
\hline
\end{oldtabular}}}
\hfill
\subfloat[MZNC 2014, Fixed category with \solver and MinisatID.]{
\centering
\label{tab:fixed14}
\scalebox{0.9}{
\begin{oldtabular}{cc}
\hline
Solver & Score \\
\hline

\textit{\textbf{sunny-cp-seq-pre}} & \textit{\textbf{835.44}}\\
\textit{Chuffed} & \textit{831.32}\\
\textit{\textbf{sunny-cp-seq}}  & \textit{\textbf{763.55}}\\
Opturion CPX  & 621.73\\
OR-Tools  & 620.34\\
SICStus Prolog  & 555.61\\
Choco  & 503.29\\
MinisatID & 472.90\\
\textit{Gecode}  & \textit{482.61}\\
\textit{G12/LazyFD}  & \textit{434.81}\\
JaCoP  & 405.66\\
\textit{G12/FD}  & \textit{293.21}\\
Picat CP  & 291.53\\

\hline
\end{oldtabular}}}
\caption{Results of MZNCs 2013--2014. 
Portfolio solvers are in bold font, parallel solvers are marked with *, not eligible 
solvers are in italics.\label{tab:results1314}}
\end{table}
The first portfolio solver that attended a MiniZinc Challenge in 2013 (see Table \ref{tab:open13}) 
was based on Numberjack platform~\cite{numberjack}. In the following years, 
\solver was unfortunately the only portfolio solver that entered the 
competition.


In 2014, \solver was a sequential solver running just one solver at time. We will 
denote it with \solverseq to distinguish such version from 
the current parallel one.
\solverseq came with two versions: 
the default one and a version with 
pre-solving denoted in Table \ref{tab:open14} as 
\texttt{sunny-cp-seq-pre}. In the latter a 
static 
selection of solvers was run for a short time, before executing 
the default version in the remaining time. Both of the two versions used 
the same portfolio of 7 solvers, viz. Chuffed, CPX, G12/FD, G12/LazyFD,
Gecode, MinisatID, MZN/Gurobi.
For more details, we refer the reader to \citeNS{sunnycp}.

\solverseq improved on Numberjack and obtained respectable results: the two variants ranked 
4\textsuperscript{th} and 7\textsuperscript{th} out of 18. 
\solverseq had to compete also with parallel solvers and all its solvers except 
MinisatID and MZN/Gurobi adopted the ``fixed'' strategy, i.e., they used the 
search heuristic defined in the problems instead of their preferred 
strategy. As described by \citeNS{sunny_rcpsp}, we realised 
afterward that this choice is often not rewarding.

To give a measure of comparison, as shown in Table 
\ref{tab:fixed14}, \solverseq in the ``Fixed''\footnote{According to 
MZNC rules, each solver in the Fixed category that has not a Free version 
is automatically promoted in the Free category (analogously, solvers in 
the Free category can be entered in the Parallel category, and then 
in turn in the Open category).
} 
category---where sequential solvers must 
follow the search heuristic defined in the model---would have been ranked 
1\textsuperscript{st} and 3\textsuperscript{rd}.
Moreover, unlike other competitors,  
the results of \solverseq were computed by including also the MiniZinc-to-FlatZinc\footnote{MZNC uses the MiniZinc language to specify the problems, while  
the solvers use the lower level specification language FlatZinc, which 
is obtained by compilation from MiniZinc models.} conversion time 
since, by its nature, \solver can not be a FlatZinc solver (see 
\citeNS{sunnycp} 
for more details).
This penalised \solverseq especially for the easier instances. 

\subsection{MiniZinc Challenge 2015}
\label{sec:mznc15}
Several enhancements of \solverseq were implemented after the MZNC 2014:  
\textit{(i)} \solver became parallel, enabling the simultaneous execution of its solvers 
while retaining the bounds communication for COPs; \textit{(ii)} new 
state-of-the-art solv\-ers were incorporated in its portfolio; \textit{(iii)} 
\solver became more stable, usable, configurable and flexible. 
These improvements, detailed by \citeNS{sunnycp2} where \solver has been tested on large 
benchmarks, have been reflected in its 
performance in the MZNC 2015. 

\solver participated in the competition with two versions: a default 
one and an ``eligible'' one, denoted \solverl in Table \ref{tab:results15}. The 
difference is that \solverl did not include 
solvers developed by the organisers of the challenge, and therefore 
was eligible for prizes. 
\solverl used 
Choco, Gecode, HaifaCSP, iZplus, MinisatID, Opturion CPX and OR-Tools solvers, 
while \solver used also Chuffed, MZN/Gurobi, G12/FD and G12/LazyFD. Since the 
availability of eight logical cores, \solver performed algorithm selection for 
computing and distributing the SUNNY sequential schedule, 
while \solverl launched all its solvers in parallel.

\begin{table}
\centering
\subfloat[Open category.\label{tab:open15}]{
\scalebox{1}{
\begin{oldtabular}{ccc}
\hline
Solver & Score & Incomplete \\
\hline
\textit{\textbf{sunny-cp}} * & \textit{\textbf{1351.13}} & \textit{\textbf{1175.2}}\\ 
\textit{Chuffed}  & \textit{1342.37} & \textit{1118.16}\\ 
\textbf{sunny-cp}$^-$ * & \textbf{1221.88} & \textbf{1156.25}\\ 
OR-Tools *  & 1111.83 & 1071.67\\
Opturion CPX  & 1094.09 & 1036.65\\ 
\textit{Gecode} *  & \textit{1049.49} & \textit{979.05}\\ 
Choco *  & 1027.65 & 989\\ 
iZplus *  & 1021.13 & 1082.92\\ 
JaCoP  & 914.97 & 865.64\\ 
Mistral *  & 872.35 & 878.53\\ 
MinisatID  & 835.01 & 793.74\\ 
\textit{MZN/CPLEX} *  & \textit{799.92} & \textit{686.64}\\ 
\textit{MZN/Gurobi} *  & \textit{774.3} & \textit{697.12}\\ 
Picat SAT  & 744.53 & 626.61\\ 
MinisatID-MP  & 637.14 & 700.35\\ 
\textit{G12/FD}  & \textit{629.94} & \textit{664.79}\\ 
Picat CP  & 617.22 & 654.81\\ 
Concrete  & 533.42 & 657.2\\ 
YACS *  & 404.01 & 553.51\\ 
OscaR/CBLS  & 403.61 & 536.17\\
\hline
\end{oldtabular}}}
\hfill
\subfloat[Free Category with \solver.\label{tab:free15}]{
\scalebox{1}{
\begin{oldtabular}{ccc}
\hline
Solver & Score & Incomplete \\
\hline
\textit{\textbf{sunny-cp}} *  & \textit{\textbf{1423.58}} & \textit{\textbf{1256.65}}\\
\textit{Chuffed}  & \textit{1387.95} & \textit{1166.56}\\
\textbf{sunny-cp}$^-$ *  & \textbf{1304.39} & \textbf{1240.88}\\
Opturion CPX  & 1146.18 & 1091.76\\
iZplus  & 1070.15 & 1093.26\\
OR-Tools  & 994.41 & 917.17\\
Mistral  & 960.16 & 937.01\\
JaCoP-fd  & 912.1 & 838.77\\
\textit{Gecode}  & \textit{908.32} & \textit{867.82}\\
Choco  & 864.39 & 887.08\\
MinisatID  & 828.2 & 791.23\\
\textit{MZN/CPLEX}  & \textit{786.11} & \textit{698.77}\\
Picat SAT  & 780.13 & 709.62\\
\textit{MZN/Gurobi}  & \textit{724.27} & \textit{654.7}\\
MinisatID-MP  & 623.58 & 688.47\\
Picat CP  & 618.78 & 633.61\\
\textit{G12/FD}  & \textit{589.65} & \textit{607.02}\\
Concrete  & 560.16 & 676.08\\
YACS  & 458.81 & 601.81\\
OscaR/CBLS  & 418.67 & 539.73\\
\hline
\end{oldtabular}}}
\caption{Results of MZNC 2015. 
Portfolio solvers are in bold font, parallel solvers are marked with *, not eligible 
solvers are in italics.\label{tab:results15}}
\end{table}
Table \ref{tab:results15} shows that \solver is the 
overall best solver while \solverl won the 
gold medal since Chuffed---the best sequential solver---was not eligible for 
prizes. 
The column ``Incomplete'' refers to the MZNC 
score computed without giving any point for proving optimality or 
infeasibility. This score, meant to evaluate local search solvers, only takes 
into account the quality of a solution. Note that with this metric 
also \solverl overcomes Chuffed, without having it in the portfolio. 
 
Several reasons justify the success of \solver in MZNC 2015. Surely the 
parallelisation on multiple cores of state-of-the-art solvers was decisive, 
especially because it was cooperative thanks to 
bounds sharing mechanism.
Moreover, differently from MZNC 2014, all the 
solvers were run with their free version instead of the fixed one.
Furthermore, the MZNC rules were less penalising for portfolio solvers since for 
the first time in the history of the MZNCs  
the total solving time included also the MiniZinc-to-FlatZinc conversion time.

We underline that the constituent solvers of \solver 
do not exploit multi-threading. 
Hence, the parallel solvers marked with * in Table \ref{tab:open15} 
are not the constituent solvers of \solver but their (new) parallel 
variants. 


The overall best single solver is Chuffed, which 
is sequential. Having it in the 
portfolio is clearly a benefit for 
\solver. However, even without Chuffed, \solverl is able to provide solutions of 
high quality (see ``Incomplete'' column of 
Table \ref{tab:results15}) proving that also the other solvers are important for 
the success of \solver. 
We remark that---as pointed out also by \citeNS{sunnycp}---when compared
to the best solver for a given problem, a portfolio solver 
always has additional overheads 
(e.g., due to feature extraction or memory contention issues) that penalise 
its score.

The 100 problems of MZNC 2015 are divided into 20 
different problem classes, each of which 
consisting of 5 instances: in total, 
10 CSPs and 90 COPs.
\solver was the best solver for only two classes: \texttt{cvrp} and
\texttt{freepizza}.
Interestingly, for the whole \texttt{radiation} problem class,  
\solverl scored 0 points because it always provided an 
unsound answer due to a buggy solver. 
This is a sensitive issue that should not be overlooked. On the one hand, a 
buggy solver 
inevitably affects the whole portfolio making it buggy as well. On the other 
hand, not using an unstable solver may penalize the global performance since 
experimental solvers like Chuffed and iZplus 
can be very efficient even if not yet in a 
stable stage.

As we shall see also in Section \ref{sec:mznc16},
unlike SAT but similarly to SMT field, most CP solvers 
are not fully reliable (e.g., in MZNC 2014 one third of the solvers provided 
at least an unsound answer).
When unreliable solvers are used, a possible way to mitigate the problem 
is 
to verify \emph{a posteriori} the solution. For instance, another constituent 
solver can be used for double-checking a solution. Obviously, checking all the 
solutions of all the solvers implies a slowdown in the solving time.
Note that the biggest 
problems arise when the solver does not produce a solution or when it declares 
a sub-optimal solution as optimal. In the first case, since solvers 
usually do not present a proof of the unsatisfiability, checking the 
correctness of the answer requires solving the same problem from scratch. In 
the second case, the presence of a solution may simplify the check of the 
answer, but checking if a solution is optimal is still an NP-hard problem.

In MZNC 2015 \solver checked 
HaifaCSP, since its author warned us about its unreliability.
This allowed \solver to detect 21 incorrect answers. Without this check its
performance would have been 
dramatically worse: \solver would have scored 87.5 points less---thus 
resulting 
worse than Chuffed---while \solverl would have scored 206.84 points less, passing 
from the gold 
medal to no medal.
However, this check was not enough: due to bugs in other constituent 
solvers 
\solver provided 5 wrong 
answers, while \solverl provided 7 wrong answers.

\subsection{MiniZinc Challenge 2016}
\label{sec:mznc16}

In the MiniZinc Challenge 2016 we enrolled three versions, namely: 
\solver, \solverl, and \solverll.

\solver was not eligible for prizes and added to the portfolio of MZNC 2015 
three new solvers: JaCoP, Mistral, and Picat-SAT.

\solverl contained only the eligible solvers of \solver, i.e., 
Choco, Gecode, HaifaCSP, JaCoP, iZplus, MinisatID, Mistral, Opturion CPX, OR-Tools, 
Picat.\footnote{We did not have an updated version of Choco and Opturion 
solvers, so we used their 
2015 version.}

\solverll contained only the solvers of \solverl that never won a medal 
in the Free category of the last three challenges, i.e., Gecode, HaifaCSP, JaCoP, MinisatID, Mistral, Picat. 

Ideally, we aimed to measure the contribution of the supposedly best solvers of \solverl.
Conversely, to \solver and \solverl, \solverll does not need to schedule its 
solvers, having fewer solvers than 
available cores. It just launches all its solvers in parallel.




\begin{table}
\centering
\subfloat[Open category.\label{tab:open16}]{
\scalebox{1}{
\begin{oldtabular}{ccc}
\hline
Solver & Score & Incomplete \\
\hline
\textit{LCG-Glucose} & \textit{1899.23} & \textit{1548.2}\\
\textit{\textbf{sunny-cp}} * & \textit{\textbf{1877.79}} & \textit{\textbf{1616.19}}\\
\textit{Chuffed} & \textit{1795.57} & \textit{1486.8}\\
\textit{LCG-Glucose-UC} & \textit{1671.52} & \textit{1306.26}\\
\textbf{sunny-cp}$^{--}$ * & \textbf{1620.82} & \textbf{1486.11}\\
\textit{MZN/Gurobi} * & \textit{1499.04} & \textit{1308.18}\\
HaifaCSP & 1448.35 & 1343.54\\
\textit{MZN/CPLEX} * & \textit{1436.05} & \textit{1287.09}\\
Picat SAT & 1423.81 & 1336.36\\
iZplus * & 1374.12 & 1446.36\\
\textbf{sunny-cp}$^-$ * & \textbf{1365.31} & \textbf{1205.73}\\
Choco * & 1342.41 & 1390.21\\
OR-Tools * & 1115.8 & 1258.51\\
\textit{Gecode} * & \textit{1110.19} & \textit{1137.21}\\
MinisatID * & 992.12 & 1002.17\\
\textit{MZN/SCIP} & \textit{985.37} & \textit{1011.25}\\
JaCoP & 923.78 & 1041.03\\
Mistral * & 826.61 & 935.8\\
\textit{MZN/CBC} & \textit{754.77} & \textit{827.06}\\
SICStus Prolog & 754.33 & 837.57\\
G12/FD & 703.14 & 829.39\\
Concrete & 583.9 & 627.36\\
Picat CP & 475.47 & 651.63\\
OscaR/CBLS & 468.5 & 708\\
Yuck * & 316 & 412\\
\hline
\end{oldtabular}}}
\hfill
\subfloat[Open Category without the instances on which \solverl failed.\label{tab:nobug16}]{
\scalebox{1}{
\begin{oldtabular}{ccc}
\hline
Solver & Score & Incomplete \\
\hline
\textit{\textbf{sunny-cp}} *  & \textit{\textbf{1054.83}} & \textit{\textbf{928.95}}\\
\textit{LCG-Glucose}  & \textit{1029.43} & \textit{876.56}\\
\textit{Chuffed}  & \textit{993.79} & \textit{844.42}\\
\textbf{sunny-cp}$^{-}$ *  & \textbf{982.7} & \textbf{893.39}\\
\textit{LCG-Glucose-UC}  & \textit{929.28} & \textit{748.17}\\
\textbf{sunny-cp}$^{--}$ *  & \textbf{899.47} & \textbf{875.6}\\
\textit{MZN/Gurobi} *  & \textit{862.26} & \textit{705.18}\\
\textit{MZN/CPLEX} *  & \textit{829.12} & \textit{704.59}\\
iZplus *  & 779.88 & 778.98\\
HaifaCSP  & 777.91 & 775.48\\
Picat SAT  & 735.82 & 713.71\\
Choco *  & 700.46 & 765.13\\
Gecode *  & 633 & 639.35\\
OR-Tools *  & 560.5 & 659.38\\
\textit{MZN/SCIP}  & \textit{545.85} & \textit{535.75}\\
MinisatID *  & 498.33 & 539.69\\
SICStus Prolog  & 437.33 & 510.66\\
JaCoP  & 433.76 & 555.49\\
\textit{MZN/CBC}  & \textit{421.32} & \textit{453.06}\\
Mistral *  & 382.68 & 470.87\\
\textit{G12/FD}  & \textit{374.56} & \textit{430.27}\\
Concrete  & 291.42 & 327.7\\
Picat CP  & 260.79 & 334.13\\
OscaR/CBLS  & 216.5 & 286.5\\
Yuck *  & 171 & 181\\
\hline
\end{oldtabular}}}
\caption{Open class results of MZNC 2016. 
Portfolio solvers are in bold font, parallel solvers are marked with *, not eligible 
solvers are in italics.\label{tab:results16}}
\end{table}
Table \ref{tab:open16} shows the Open category ranking of the MZNC 2016. 
These results are somehow unexpected if compared with those 
of the previous editions. For the first time, Chuffed has been outperformed by a 
sequential solver, i.e., the new, experimental LCG-Glucose---a lazy clause generation solver 
based on Glucose SAT solver. Surprisingly, solvers like OR-Tools, iZplus, Choco had subdued 
performance. Conversely, HaifaCSP and Picat-SAT performed very well. The sharp improvement 
of the solvers based on Gurobi and CPLEX is also clear, arguably due to a better linearisation of the 
MiniZinc models~\cite{linmod}. Local search solvers still appear immature.

The results of \solver are definitely unexpected. In particular, it appears quite strange that 
\solverl performed far worse than \solverll although having more, and ideally better, solvers.
We then thoroughly investigated this anomaly since, as also 
shown in Amadini et al. \citeyear{sunnycp2,sunny_rcpsp}, the dynamic scheduling 
of the available 
solvers is normally more fruitful than statically running an arbitrarily good subset of them 
over the available cores.

Firstly, we note that for the easier instances 
\solverll is inherently faster than \solver and \solverl because it does not need to schedule 
its solvers, and therefore it skips the feature extraction and the algorithm selection phases 
of the SUNNY algorithm~\cite{sunny}. Nevertheless, most of the MZNC 2016 
instances are not 
easy to solve.

Another reason is that \solverll always runs HaifaCSP and Picat-SAT, 
two solvers that performed better than expected, while \solverl executes Picat-SAT only 
for 37 problems. Nonetheless, \solverl always executes HaifaCSP so also this explanation 
can not fully explain the performance difference.

The actual reason behind the performance gap relies on some  
\textit{buggy solvers} which belongs to \solverl but not to 
\solverll.\footnote{With the term ``buggy solver'' we not necessarily mean that 
the solver itself is 
actually buggy. The problems may arise due to a misinterpretation of the 
FlatZinc 
instances or to the wrong 
decomposition of global constraints \cite{handbook}.}
In our pre-challenge tests we did not notice inconsistencies in any 
of the solvers, except for Choco. So we decided to check the solutions only for Choco and 
HaifaCSP (the latter because of the unreliability shown in the MZNC 2015, see Section \ref{sec:mznc15}).
However, none of these solvers gave an unsound outcome in the MZNC 2016. 
Conversely, Opturion and OR-Tools solvers provided a lot of incorrect, and 
unfortunately unchecked, answers.
We also noticed that for some instances our version of 
Mistral failed when restarted with a new bound, while on the same instances the Free version of 
Mistral provided a sound outcome. 

In total, \solverl gave 24 wrong 
answers,\footnote{Namely, all the 5 instances
of \texttt{depot-placement}, \texttt{gfd-schedule}, and \texttt{nfc} classes; 4 instances of \texttt{tpp} class; 
1 instance of \texttt{cryptanalysis}, \texttt{filter}, \texttt{gbac}, 
\texttt{java-auto-gen}, \texttt{mapping} classes.} meaning that it 
competed only on the 76\% of the problems of MZNC 2016. \solverll failed instead on 5 instances.

Table \ref{tab:nobug16} shows the results without the 24 instances 
for which \solverl gave an incorrect answer. 
We underline that this table has a purely indicative value: for a more comprehensive 
comparison, also the instances where other solvers provided an incorrect answer 
should be removed.
On these 76 problems 
\solver overcomes LCG-Glucose, while \solverl impressively 
gains 7 positions and becomes gold medallist being the first of the eligible solvers.
\solverll however behaves well (silver medallist), being overtaken by \solverl only.

Note that the results of \solver are good also in the original ranking of Table \ref{tab:open16}
since, being this version not eligible for prizes, the organisers enabled the solutions checking of 
G12/LazyFD, HaifaCSP, Mistral, Opturion, OR-Tools. This allowed \solver to detect 19 incorrect 
answers. 

An interesting insight is given by the Incomplete score, which does not give any benefit 
when a solver concludes the search (i.e., when optimality or unsatisfiability is proven). 
As observed also in Section \ref{sec:mznc15}, with this metric  \solver can significantly 
overcome a solver that has a greater score (e.g., see the Incomplete 
\solver in Table \ref{tab:open16}). This confirms the attitude of \solver in finding 
good solutions even when it does not conclude the search.

\section{Conclusions}
\label{sec:conc}

We presented an overview of \solver, a fairly recent CP portfolio solver 
relying on the SUNNY algorithm, and we discussed its performance in 
the MiniZinc Challenge---the annual international competition for 
CP solvers. 

In the MiniZinc Challenge 2014 \solver received an honourable mention, 
in 2015 it has been the first portfolio solver to win 
a (gold) medal, and in 2016---despite several issues with buggy solvers---it 
confirmed the first position.

For the future of CP portfolio solvers, it would be interesting 
having more portfolio competitors to improve the state of the art in this field.
Different portfolio approaches have been already compared w.r.t. \solver and its 
versions \cite{sunny,paper_amai,cilc_2015,sunny_plus}. 

The Algorithm Selection approaches of the ICON 
Challenge 2015~\cite{aslib_chall} might be adapted to deal with generic CP 
problems. 
The SUNNY algorithm itself, which is competitive in the CP scenarios of 
\citeS{sunny,paper_amai},\footnote{
We submitted such scenarios, namely \texttt{CSP-MZN-2013} and 
\texttt{COP-MZN-2013},  
to the Algorithm Selection Library~\cite{aslib}.
}
provided very poor 
performance in the SAT scenarios of the ICON Challenge and 
\citeNS{sunny_plus} show that it can be strongly improved with a proper 
training phase.


\solver runs in 
parallel different single-threaded solvers. This choice so far has proved to be 
more 
fruitful than parallelising the search of a single solver. However, the 
possibility of using multi-threaded solvers may have some benefits when solving 
hard problems as shown by \citeNS{3s-par} for SAT problems.

The multi-threaded execution also enables search splitting strategies.
It is not clear to us if the use of all the available cores, as done by 
\solver, is the best possible strategy. As shown by  
\citeNS{parallelSatInsights} it is possible that running in parallel all the 
solvers on the same multicore machine slows down the execution of the 
individual solvers. 
Therefore, it may be more convenient to leave free one or more cores 
and run just the most promising solvers.
Unfortunately, it is hard to extrapolate a general pattern to understand 
the interplay between the solvers and their resource consumption.

One direction for further investigations, clearly emerged from the challenge 
outcomes, concerns how to deal with  
unstable solvers. Under these circumstances it is important to find a trade-off 
between reliability and performance. Developing an automated way of checking a 
CP solver outcome when the answer is ``unsatisfiable problem'' or 
``optimal solution'' is not a trivial challenge: we can not merely do a solution 
check, but we have to know and check the actual \textit{explanation} for which the 
solver provided such an outcome.

A major advancement for CP portfolio solvers would be having API for injecting 
constraints at runtime, without stopping a running solver. Indeed, 
interrupting a solver means losing all the knowledge it has collected so far. 
This is particularly bad for Lazy Clause Generation solvers, and in general for 
every solver relying on no-good learning.

Another interesting direction for further studies is to consider the impact of 
the global constraints \cite{handbook} on the performances of the portfolio 
solver. It is well-known that the propagation algorithms and the 
decompositions used for global constraints are the keys of solvers 
effectiveness. We believe that the use of solvers supporting different global 
constraint decompositions may be beneficial.

We underline that---even if focused on Constraint Programming---this work can be 
extended to other fields, e.g., Constraint Logic Programming, Answer-Set Programming 
or Planning, where portfolio solving has been used only marginally.

To conclude, in order to follow the good practice 
of making the tools publicly available and easy to 
install and use, we stress that \solver is 
publicly available at \url{https://github.com/CP-Unibo/sunny-cp} and can be 
easily installed, possibly relying on the Docker container technology for avoiding 
the installation of its constituent solvers. All the results 
of this paper can be reproduced and verified by using the web interface of 
\url{http://www.minizinc.org/challenge.html}.




%

\section*{Acknowledgements}
We are grateful to all the authors and developers of the constituent solvers of \solver, 
for providing us the tools and the instructions to use the solvers. 

We 
thank all the MiniZinc Challenge staff, and in particular Andreas Schutt, for the 
availability and technical support.

This work was supported by the EU project FP7-644298 {HyVar:
Scalable
Hybrid Variability for Distributed, Evolving Software
Systems}.

\bibliographystyle{acmtrans}
\bibliography{biblio}

\end{document}